\documentclass[accepted]{uai2026} 

\usepackage[american]{babel}

\usepackage{natbib} 
\usepackage{mathtools} 
\usepackage{booktabs} 
\usepackage{tikz} 

\newcommand{\class}{c}

\newcommand{\Classes}{\mathcal{C}}

\newcommand{\prediction}{\hat{\class}}

\newcommand{\classifier}{h}

\newcommand{\feature}[1]{f_{#1}}

\newcommand{\FeatureSet}[1]{\mathcal{F}_{#1}}
\newcommand{\features}{f}

\newcommand{\FeaturesSet}{\mathcal{F}}
\newcommand{\numfeatures}{N}

\newcommand{\prob}{P}

\newcommand{\learneddistr}{\prob_{\mathrm{classif}}}
\newcommand{\testdistr}{\prob_{\mathrm{test}}}
\newcommand{\traindistr}{\prob_{\mathrm{train}}}

\newcommand{\testset}{D_{\mathrm{test}}}
\newcommand{\trainset}{D_{\mathrm{train}}}

\newcommand{\maxprob}{u_{\mathrm{max}}}
\newcommand{\margin}{u_{\mathrm{conf}}}
\newcommand{\entropy}{u_H}
\newcommand{\aleatunc}{u_{a}}
\newcommand{\totunc}{u_{t}}
\newcommand{\epistunc}{u_{e}}
\newcommand{\Nmodels}{M}

\newcommand{\noise}{\beta}

\newcommand{\robglob}{r_{\mathrm{glob}}}
\newcommand{\robloc}{r_{\mathrm{loc}}}

\definecolor{darkgreen}{rgb}{0.01, 0.65, 0.18}
\definecolor{rob_blue}{rgb}{0.0, 0.75, 1.0}
\definecolor{unc_yellow}{rgb}{0.95, 0.7, 0.15}
\definecolor{hybrid_green}{rgb}{0.1, 0.7, 0.2}

\title{Robustness Quantification and Uncertainty Quantification: \\Comparing Two Methods for Assessing the Reliability of Classifier Predictions}

\author[1]{Adri\'an~Detavernier}
\author[1]{Jasper~De~Bock}
\affil[1]{%
  Foundations Lab for imprecise probabilities\\
  Ghent University\\
  Belgium
}

\begin{document}
\maketitle

\begin{abstract}
    We consider two approaches for assessing the reliability of the individual predictions of a classifier: Robustness Quantification (RQ) and Uncertainty Quantification (UQ).
    We explain the conceptual differences between the two approaches, compare both approaches on a number of benchmark datasets and show that RQ is capable of outperforming UQ, both in a standard setting and in the presence of distribution shift.
    Beside showing that RQ can be competitive with UQ, we also demonstrate the complementarity of RQ and UQ by showing that a combination of both approaches can lead to even better reliability assessments.
\end{abstract}

\section{Introduction}
\label{sec:intro}
Due to its vast capabilities, AI has become ubiquitous, its use cases ranging from automating simple tasks to making decisions in high-risk settings.
In some cases, especially the ones where the stakes are high, we are not only interested in the overall performance of the model, but also in the quality or, to be more precise, the reliability of each single prediction.
If your own health is at stake for instance, you care less about how well the model performs on average; you only want to know whether you can rely on the model's prediction in your particular case.
So, in an ideal world, we'd want to know for each prediction of an AI model how reliable it is.
For the least reliable predictions, a second opinion of an expert could then be asked, more data could be collected, etc.

One of the more popular applications of AI models, and the one we focus on in this paper, is classification.
In that case, the goal of the model is to predict the correct class \(\class\) of a given instance out of a set of possible classes \(\Classes\).
An instance is then usually described using a number of features (\(\numfeatures\) in total). The value \(\feature{i}\) of the \(i\)-th feature takes values in a set \(\FeatureSet{i}\), which we take to be finite because we'll restrict ourselves to discrete features. We'll call the vector \(\features \coloneq (\feature{1}, \dots, \feature{\numfeatures})\) the (set of) features of said instance, which takes values in \(\FeaturesSet \coloneq \FeatureSet{i} \times \dots \times \FeatureSet{\numfeatures}\).
In practice a classifier then, given an instance (e.g. a patient), uses its features \(\features\) (e.g. the patient's medical data) to try to predict the correct class \(\class\) (e.g. the condition/disease/treatment of the patient).
We'll denote the class predicted by the classifier as \(\prediction\).

For each such prediction \(\prediction\) of a classifier, we can now try to assess how reliable it is.
In this work, we consider two methods for doing so, namely \emph{uncertainty quantification} \citep{hullerenwillem,sale2024label} and \emph{robustness quantification} \citep{NIPS2014_09662890,pmlr-v62-mauá17a,correia2020robustclassificationdeepgenerative,detavernier2025robustness}.
Since both approaches aim to quantify (an aspect of) the reliability of individual predictions, these approaches can be put under a common umbrella, which we will refer to as \emph{reliability quantification}, as depicted in the figure below.

\begin{figure}[!h]
    \centering
    \begin{tikzpicture}[fill=gray, yscale=.8]
        \fill[hybrid_green, opacity=.2, rounded corners=5pt] (-2,-1.5) rectangle (3,1.35);
        \fill[draw=black, fill=none, rounded corners=5pt, line width=.3pt] (-2,-1.5) rectangle (3,1.35);
        \fill[unc_yellow, opacity=.7] (-.3,0.1) circle (1);
        \fill[rob_blue, opacity=.7] (1.3,0.1) circle (1) (0.8,-1);
        \draw (-.3,0.1) circle (1) (0.3,-.9)  node [unc_yellow, below left] {Uncertainty Q.}
        (1.3,0.1) circle (1) (0.8,-.9)  node [rob_blue, below right] {Robustness Q.} (.5,1.35) node [text=hybrid_green, above] {Reliability Quantification};
        \begin{scope}
            \clip (-.3,0.1) circle (1);
            \fill[hybrid_green, opacity=0.3] (1.3,0.1) circle (1) (0.8,-1);
        \end{scope}
    \end{tikzpicture}
    \caption{Situating robustness quantification and uncertainty quantification with respect to reliability quantification.}\label{fig:RelQ}
\end{figure}
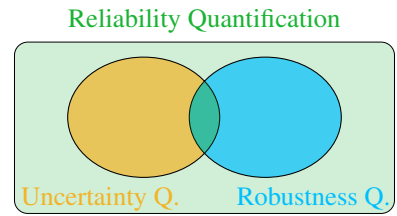

What both approaches also have in common is that they are based on the core idea that there is a lot of uncertainty involved when learning a model from data and using it to make predictions, and that this uncertainty is one of the main reasons why such predictions can be unreliable.
Uncertainty quantification tries to quantify (or perhaps we should say estimate) this uncertainty, for the predictions associated with individual instances, in the form of numerical uncertainty measures.
Robustness quantification, on the other hand, tries to quantify the amount of (epistemic) uncertainty the model could handle without changing its prediction, regardless of how much uncertainty there actually is, making it more suitable for situations in which estimating uncertainty is hard.
In the former approach, we expect low uncertainty to correspond with reliable predictions; in the latter approach we expect hight robustness to indicate reliable predictions.

In a side by side comparison, it was recently demonstrated for the Naive Bayes Classifier (NBC) that robustness quantification is very competitive with uncertainty Quantification when it comes to assessing the reliability of the predictions of a classifier, at least for artificial data and in the presence of distribution shift or when there was a limited amount of training data \citep{detavernier2025robustness}.
Our contribution aims to conduct a more comprehensive comparison of robustness and uncertainty quantification by benchmarking both approaches on real (instead of artificially generated) datasets, not only for the NBC, but also for Generative Forests (GeFs).
Besides conducting a side by side comparison, we also investigate their complementarity: instead of only comparing the two and studying which one is better, why not combine them?
Since, on a conceptual level, robustness and uncertainty cover different aspects of reliability, it seems plausible that such a combination could lead to even better reliability assessments.
In our experiments on benchmark datasets, we show that this is indeed the case.

\section{Uncertainty and Robustness for Probabilistic Generative Classifiers}
\label{sec:UqandRQ}
Formally, a classifier \(\classifier: \FeaturesSet \to \Classes\) is a function from the set of all possible sets of features to the set of possible classes.
The uncertainty and robustness measures we consider in this work are designed for probabilistic classifiers and, in the case of the robustness measures, generative ones. Since our features are discrete, such a probabilistic generative classifier is completely determined by a probability mass function \(\learneddistr\) on \(\Classes\times\FeaturesSet\).
For a given set of features \(\features\), the predicted class \(\prediction\) is then typically (assuming 0-1 loss) the one with the highest probability given the features:
\begin{equation*}
	\prediction \coloneq \classifier(\features) = \arg \max_{\class \in \Classes} \learneddistr(\class \vert \features),
\end{equation*}
where \(\learneddistr(\cdot \vert \features)\) is obtained from \(\learneddistr\) through Bayes' rule.
In our experiments further on, we make use of two types of probabilistic generative classifiers:
the Naive Bayes Classifier (NBC) \citep{naiveBayes} that assumes the features to be independent given the class,
and Generative Forests (GeFs) \citep{NEURIPS2020_8396b14c}, a subclass of Probabilistic Circuits that is closely related to Decision Trees and Random Forests.

In practice, (the probability mass function of) a generative classifier \(\learneddistr\) is learned using a training set \(\trainset\) of (correctly) labeled instances, and its performance is then evaluated on a different set of instances, called the test set \(\testset\).
A common assumption, which we will follow here, is that these datasets are sampled from underlying distributions, whose probability mass functions we will denote by \(\traindistr\) and \(\testdistr\), respectively.
An additional assumption that is often made is that these two distributions are the same, that is, \(\traindistr = \testdistr\). This need not be the case in practice though, for example when a model is trained on one type of data and then deployed in a different setting. This phenomenon is called \emph{distribution shift}, and is something that we will study the effect of in some of our experiments.

The ideal classifier is the one for which \(\learneddistr = \testdistr\).
However, even in this ideal case, the accuracy of the predictions issued by this classifier will typically not be 100\%, meaning that even then there still is uncertainty present in the prediction.
This has to do with the intrinsic variability present in the task at hand: two instances with the same features could in practice have a different class, either due to inherent randomness or because not enough information is captured in the set of features to distinguish these cases.
This remaining uncertainty is completely captured by \(\testdistr(\cdot \vert \features)\), and we will refer to it as \emph{aleatoric uncertainty}.

Unfortunately, the case where the classifier perfectly learns \(\testdistr\) is unrealistic.
The more realistic scenario is that the learned classifier differs from the ideal one, that is, \(\learneddistr \neq \testdistr\).
The fact that \(\learneddistr\) and \(\testdistr\) need not be the same, is a completely different type of uncertainty associated with classification, which we call \emph{epistemic uncertainty}.
Possible sources of this uncertainty are structural modelling assumptions (such as the independence assumption of an NBC), the fact that \(\learneddistr\) is based on a finite (and hence possibly too small or unrepresentative) training set, or the presence of distribution shift. Of these three sources, the first two lead to \(\traindistr \neq \learneddistr\) whereas distribution shift leads to \(\traindistr \neq \testdistr\).

\subsection{Uncertainty Quantification}

Uncertainty quantification tries to quantify (or estimate) aleatoric and/or epistemic uncertainty, in the form of a numerical uncertainty measure. Many such measures have been developed, of which we consider the following six in our experiments.

A first intuitive uncertainty measure is the probability of the predicted class being wrong, at least according to the learned distribution \(\learneddistr\) given \(\features\), as defined by
\begin{equation}
    \maxprob(\features)= 1 - \learneddistr(\prediction \vert \features).
\end{equation}
In the ideal case where \(\learneddistr=\testdistr\), this would be equivalent to the probability of making a wrong decision.
It thus can be seen as an estimate of the aleatoric uncertainty for the prediction associated with \(\features\).
A second intuitive measure is the so-called margin of confidence, which is simply the probability difference between the most and second most probable class according to the learned distribution given \(\features\):
\begin{equation}
    \margin(\features) = \learneddistr(\prediction \vert \features) - \max_{\class \in \Classes \backslash \prediction}\learneddistr(\class \vert \features).
\end{equation}
Another attempt at estimating the aleatoric uncertainty of an instance with features \(\features\) makes use of the (Shannon) entropy.
This measure is the entropy of \(\learneddistr(\cdot\vert\features)\), defined by
\begin{equation}
    \entropy(\features) = -\sum_{\class \in \Classes} \learneddistr(\class\vert\features)\log_{2}\learneddistr(\class\vert\features).
\end{equation}
The motivation for using entropy as an estimate of uncertainty is because it reflects the ``peakedness'' of the distribution: the higher the entropy, the closer to uniform the distribution is.
Since a uniform distribution corresponds to not being sure at all about the prediction, a higher entropy clearly should correspond to more uncertainty.
The remaining three uncertainty measures are based on an information-theoretic decomposition of uncertainty in total, aleatoric and epistemic uncertainty \citep{pmlr-v80-depeweg18a}. Each of them is estimated by combining entropy with ensemble techniques \citep{pmlr-v80-depeweg18a,Mobiny2021,Shaker2020}.
We denote the number of trained models in each ensemble by \(\Nmodels\)---which we set to $10$ in our experiments---and use \(\prob_i\), with \(i \in \{1, \dots,\Nmodels\}\), to denote the distribution of the \(i\)-th model in the ensemble.
Each model in the ensemble is trained on a bootstrap sample (of the same size as \(\trainset\)) we obtained by sampling with replacement from the training set \(\trainset\) itself.
Total uncertainty is then estimated as the entropy of the average predicted conditional distribution, given by
\begin{equation}
    \totunc(\features) = -\sum_{\class \in \Classes}\prob_{\mathrm{av}}(\class\vert\features)\log_2\prob_{\mathrm{av}}(\class\vert\features),
\end{equation}
where \(\prob_{av}(\class\vert\features) = \frac{1}{\Nmodels}\sum_{i=1}^{\Nmodels}\prob_i(\class\vert\features)\).
Aleatoric uncertainty is the average of the entropies of the predicted conditional distributions of the models in the ensemble:
\begin{equation}
    \aleatunc(\features) = -\frac{1}{\Nmodels}\sum_{i=1}^{\Nmodels} \sum_{\class \in \Classes} \prob_i(\class\vert\features)\log_{2}\prob_i(\class\vert\features).
\end{equation}
Finally, the epistemic uncertainty is calculated as the difference between total and aleatoric uncertainty:
\begin{equation}
    \epistunc(\features) = \totunc(\features) - \aleatunc(\features).
\end{equation}
For a more in-depth discussion of these information-theoretic measures, we refer to the works of \citet{Mobiny2021} and \citet{Shaker2020}.

\subsection{Robustness Quantification}
In contrast to Uncertainty Quantification, Robustness quantification does not aim to quantify (or estimate) uncertainty. Instead, it tries to numerically quantify how much (epistemic) uncertainty the model \emph{could} handle without changing its prediction, thereby ignoring how much uncertainty there actually is.
More concretely, instead of only looking at what class is predicted by \(\learneddistr\), this approach considers neighborhoods of distributions around (the joint distribution) \(\learneddistr\).
If all distributions in such a neighborhood predict the same class as the one predicted by \(\learneddistr\), we call this prediction \emph{robust} w.r.t. said neighborhood.
If the prediction is robust, then the size of the neighborhood can be seen as a lower bound on the amount of epistemic uncertainty we can allow without changing the prediction.
By controlling the size of the neighborhood in a parametrized manner, we can increase this size until the prediction of the model is no longer robust, or thus until at least one distribution in the neighborhood predicts a different class.
The parameter value at which this happens can then be used as a robustness measure, indicating the amount of epistemic uncertainty the model could handle without changing this particular prediction.
Quantifying robustness this way has been successfully tried several times already in the literature, for different types of classifiers \citep{NIPS2014_09662890,pmlr-v62-mauá17a,correia2020robustclassificationdeepgenerative,detavernier2025robustness}. Due to their use of neighborhoods (and hence sets) of distributions, these approaches often rely on algorithmic techniques from the field of imprecise probability theory~\citep{augustin2014introduction}.

There are of course numerous types of neighborhoods that can be considered, and hence many different corresponding robustness measures.
We restrict ourselves to two (types of) such families, and thus to two (types of) robustness measures.
The first robustness measure can be applied to any probabilistic generative classifier; it considers neighborhoods of the learned (global) joint distribution \(\learneddistr\) obtained by \(\epsilon\)-contaminating~\citep{Huber1992} the latter.
We will refer to this robustness measure as the global one, since it perturbs the global joint distribution, and denote is as \(\robglob\).
Even though it might at first sigt seem very troublesome and computationally expensive to evaluate the predictions of all the distributions in such a neighborhood, it turns out that calculating this global robustness measure can in fact be reduced to evaluating the following closed-form expression:
\begin{equation}
    \robglob(\features) = \frac{\Delta}{1+\Delta},
\end{equation}
with \(\Delta = \learneddistr(\prediction, \features) - \max_{\class \in \Classes \backslash \prediction} \learneddistr(\class, \features)\) \cite{detavernier2025robustness}.
This measure can thus also be interpreted as a monotone transformation of the difference of the joint probabilities of the two most probable classes according to $\learneddistr$. In that sense, it is similar in style to the margin of confidence measure that is used in uncertainty quantification. A crucial difference, however, is that the margin of confidence is a difference of conditional probabilities, whereas the global robustness measure is a difference of joint probabilities.

A second (type of) robustness measure, which we'll denote by \(\robloc\), is tailor-made for each of the models used --- in our case either NBCs or GeFs --- by perturbing their local parameters.
In this way, we obtain neighborhoods of the learned model \(\learneddistr\) whose elements all have the same model architecture or structural assumptions as \(\learneddistr\), but differ in their choice of local parameters.
Similarly to what is done for the global robustness measure, perturbations are again obtained by \(\epsilon\)-contamination; instead of contaminating $\learneddistr$ directly, we now contaminate its local parameters.
There is no closed-form expression for \(\robloc\); computing it typically requires a combination of optimisation and binary search.
For more details about this local robustness measure, including how to efficiently compute it, we refer to the recent work of \citet{detavernier2025robustness} for the NBC case and that of \citet{correia2020robustclassificationdeepgenerative} for GeFs.

Finally, since `robustness' refers to many different concepts within ML, it is worth pointing out that robustness quantification is instance-based, meaning that it assesses the robustness of individual predictions. This sets it apart from a plethora of approaches that consider the robustness of a classifier as a whole, such as adversarial robustness \citep{bai2021recent,carlini2019evaluating} (which is sometimes instance-based but focussed on perturbations of continuous features rather than perturbations of the model), robustness against distribution shift \citep{NEURIPS2020_d8330f85}, or robust optimization \citep{ben2009robust}.

\subsection{Their Conceptual Difference}

Having introduced more precisely the concepts of uncertainty and robustness quantification, we now take a closer look at how and why (we think) these two approaches are different on a conceptual level.
For a start, the goal of the approaches differs in how the uncertainty of a prediction/model is handled: uncertainty quantification tries to quantify how much uncertainty \emph{there is}, whereas robustness quantification tries to quantify how much (epistemic) uncertainty \emph{there can be} without changing the prediction.
In the ideal scenario where there is only aleatoric uncertainty, the best possible assessment of the reliability of the prediction would therefore be \(\testdistr(\cdot\vert\features)=\learneddistr(\cdot\vert\features)\) itself, and there would then be no use for robustness quantification (nor for epistemic uncertainty quantification).
Unfortunately, this is rarely ever the case because there is often epistemic uncertainty present as well. That's when uncertainty quantification becomes more difficult, and where robustness quantification comes into play.

The difference between the two approaches is most clear when we allow for the possibility of distribution shift \(\traindistr\neq\testdistr\).
From the point of view of uncertainty quantification, we'd then like to try to quantify the amount of aleatoric uncertainty and/or epistemic uncertainty.
This task is extremely challenging, though, since we can never know how big the difference between \(\testdistr\) and \(\traindistr\) is, nor whether \(\trainset\) or \(\testset\) are representative for \(\traindistr\).
This becomes particularly clear in the perfect scenario where \(\learneddistr = \traindistr\).
In this case it is still impossible to quantify the difference between $\testdistr$ and $\learneddistr$ without access to labeled data from \(\testdistr\), and hence to quantify the epistemic uncertainty, which in its turn also makes it difficult to quantify the aleatoric uncertainty because we don't know how much of the uncertainty captured by \(\learneddistr\) is actually present in \(\testdistr\).
On the other hand, if we approach the same problem from the perspective of robustness quantification, the task does not break down, since it does not require \(\testdistr\).
We are simply interested in the extent to which $\testdistr$ can differ from $\learneddistr$ without changing the prediction, which is something that we can assess without knowing \(\testdistr\) at all.


Looking back at Figure~\ref{fig:RelQ}, we also want to point out that the intersection between uncertainty and robustness quantification is non-empty.
The reason for this is that it is possible for a certain reliability measure to have both an uncertainty- and a robustness-based interpretation. That would make it a perfect reliability measure in fact, since it would be able to capture both aspects of reliability at the same time.

\section{Evaluating Reliability Measures}\label{sec:evalmeasures}
Since uncertainty and robustness measures both share the goal of trying to assess the reliability of the individual predictions of a classifier, it makes sense to refer to both of them as \emph{reliability measures}.
Depending on the task at hand, such a measure can be used to either select the most reliable instances (for example to automate predictions for those instances) or to select the least reliable ones, and hence the hardest ones to classify (for example to classify these manually, or collect more data for them).
A perfect reliability measure would thus be able to order all instances in such a way that if we'd start rejecting instances in that order, we would first reject all wrongly classified ones, and then only at the end the correct ones.
A straightforward way of evaluating the performance of a reliability measure is therefore to look at how well it is capable of ordering a set of instances such that the misclassified instances are rejected first.

Accuracy rejection curves (ARC) offer a visual way to evaluate this \citep{pmlr-v8-nadeem10a}.
For a given reliability measure, an ARC is made by first ordering all instances in order of increasing reliability; so from high to low uncertainty for uncertainty measures, or from low to high robustness for robustness measures.
Once the order is determined, we start rejecting instances in that order, such that the ones with the lowest reliability get rejected first, and then at every step calculate the accuracy of the classifier on the remaining instances.
So, in essence, ARCs plot the accuracy as a function of the rejection rate.
Figure~\ref{fig:arc_grid}, which we'll analyse in detail later on, displays several examples of such ARCs.
Note that the higher the overall curve is for a given measure, the better the measure is performing.

Comparing (the ARCs of) different reliability measures is not always straightforward though, since which one is best can depend on the rejection rate.
This makes assigning a winner a subjective matter. To address this issue, we try to summarize the quality of such a curve in a single value. Following the approach suggested in the conclusion of the work of \citet{pmlr-v8-nadeem10a}, we use the area under the ARC, which we'll refer to as AU-ARC.
Since an ARC consists of a discrete number of points, the AU-ARC can simply be calculated by taking the average of the accuracies that correspond to all possible rejection rates.


\begin{table}
    \centering
    \caption{All datasets (UCI) used in the experiments.}\label{tab:datasets}
    \includegraphics[width=.88\linewidth]{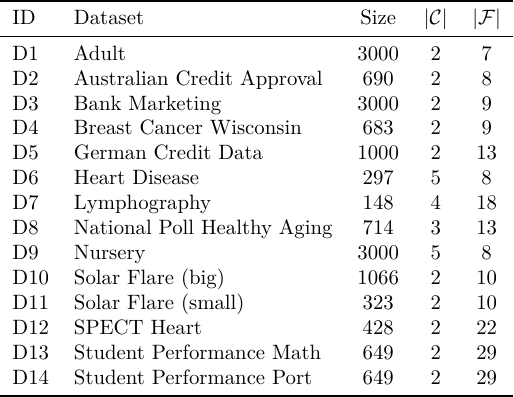}

\end{table}

To assess the performance of the different uncertainty and robustness measures we consider, we conducted experiments on (cleaned versions of) several datasets from the UCI Machine Learning Repository \citep{ucimlrepository} shown in Table~\ref{tab:datasets}.
Since we restrict ourselves to discrete features, continuous features were removed when present, as were instances with missing values.
For two particular datasets we adapted the task to one that leans more toward standard classification.
The Solar Flare dataset originally has three possible target variables that can be predicted, being the number of flares that occur, and this for three types of flares;
we turned this into a binary classification task whose aim it is to predict if at least one solar flare occurs (of any type).
Similarly, for the Student Performance dataset we predict whether the student passes or fails, instead of predicting the exact grade.
Unless the dataset provides a test set itself, we randomly split the datasets into \(\trainset\) and \(\testset\), containing 60\% and 40\% respectively, with a maximum dataset size of 3000 (training and test set combined).
For the experiments where we use the Naive Bayes Classifier, we first optimize a smoothing parameter with 5-fold cross validation on \(\trainset\).
Once the optimal smoothing parameter is found, we train the classifier on the entire training set for said parameter.
For experiments with the Generative Forests, we set the \(\mathrm{n\_estimators}\) parameter to 30 and used the given default values for the rest of them.

\section{Comparing Robustness and Uncertainty Quantification}\label{sec:comparison}
In this section we compare the performance of the robustness and uncertainty measures we discussed in Section~\ref{sec:UqandRQ} in two different settings.
The first setting is the standard one where the model is trained on the entire (unchanged) training set \(\trainset\).
In a second setting we create a scenario where the epistemic uncertainty is high(er) by limiting the number of training data and/or by introducing distribution shift.

\paragraph{Standard Setting}

The results for the standard setting are shown in Table~\ref{tab:auc_all_nbc} for the NBC and Table~\ref{tab:auc_all_gefs} for GeFs.
In the case of the NBC we see that \(\robloc\) wins 6 out of 14 times (as highlighted in bold), whereas the best uncertainty measures, namely \(\maxprob\), \(\entropy\) and \(\aleatunc\), only win 3 times.
The global robustness measure \(\robglob\) performs less well though, and does not seem competitive in this setting.
The results for the GeFs are similar but more pronounced, since \(\robloc\) wins for 7 out of 14 datasets now.
Furthermore, for several cases where local robustness wins here, it does so with a distinct margin.
The general conclusion we take from these tables is that the reliability measure that has the highest performance in most cases is the local robustness measure,
and that for GeFs in particular this difference can be very outspoken.

\begin{table}
    \centering
    \caption{The AU-ARCs for the NBC for all datasets in the standard setting.}\label{tab:auc_all_nbc}
    \includegraphics[width=\linewidth]{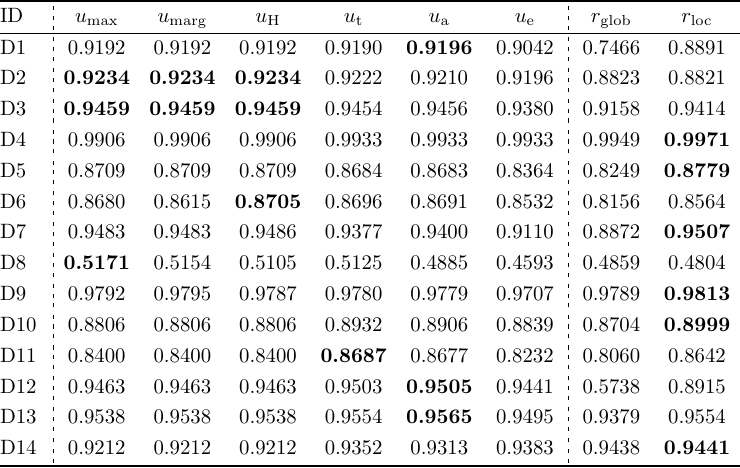}
\end{table}

\begin{table}
    \centering
    \caption{The AU-ARCs for the GeFs for all datasets in the standard setting.}\label{tab:auc_all_gefs}
    \includegraphics[width=\linewidth]{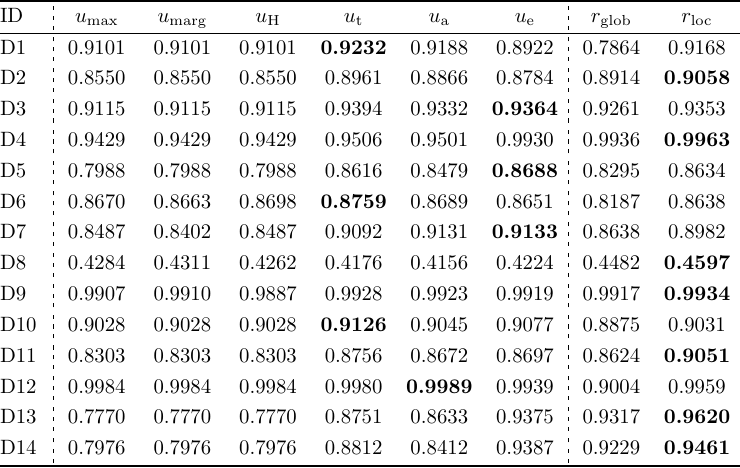}
\end{table}

\paragraph{Distribution Shift and Limited Data Setting}

\begin{figure*}[!h]
    \centering
    \includegraphics[scale=1.1, trim=8.2cm 11.3cm 2.8cm 9.6cm]{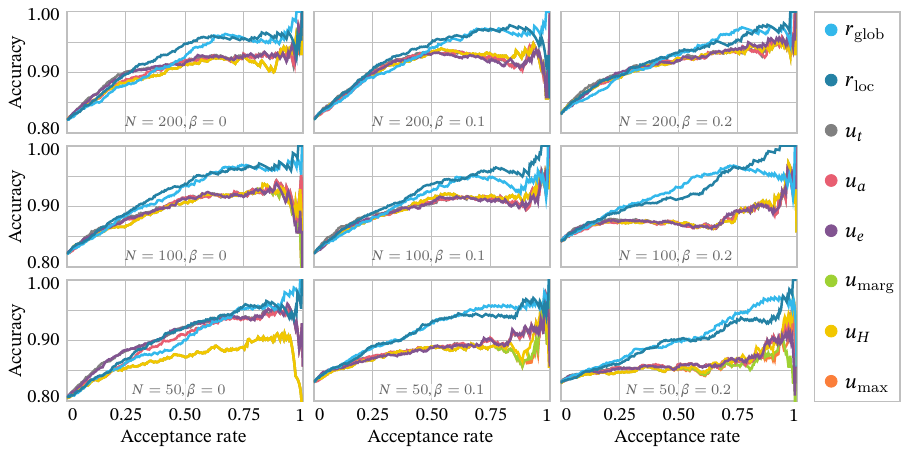}
    \caption{Each graph represents the mean of the ARCs for the NBC on the Student Performance Port (D14) dataset for a combination of training set size \(N\) and feature noise \(\noise\).}\label{fig:arc_grid}
\end{figure*}

In the second part of our side by side comparison we increase the amount of epistemic uncertainty to study how this influences the relative performance of the considered reliability measures.
Our first source of epistemic uncertainty is the training set \(\trainset\) itself.
Because it is (assumed to be) sampled from a distribution, there is always some variability to it, which can lead to it not being representative of \(\traindistr\).
The more training data we have, the less prone we are to this variability, or reversely,
by limiting the number of training, we can increase the amount of epistemic uncertainty.
In addition to limiting the number of training data, we also introduce some distribution shift, thereby increasing the amount of epistemic uncertainty even more.

In particular, to study the effect of the number of training data, we will consider different sizes \(N\) of \(\trainset\) in our experiments, with \(N \in \{50, 100, 200\}\).
To be able to control the amount of distribution shift, we opted to introduce it ourselves by corrupting the feature values of the training data.
We do this by corrupting each feature value with a probability of \(\noise\); if corrupted, we replace it by a different value that is uniformly picked from the remaining options for this feature.
The amounts of label noise we consider in our experiments are \(0.0\), \(0.10\) and \(0.20\).
For each of the 9 possible combinations of dataset size and label noise, we calculated the ARCs (and corresponding AU-ARCs) for 7 different randomly sampled \(\trainset\) and \(\testset\).
The results are shown in Figure~\ref{fig:arc_grid} for data set D14 and the NBC, where each of the 9 graphs corresponds to the average over 7 ARCs for a particular combination of \(N\) and \(\noise\).
The distribution shift \(\noise\) increases in the figure going from left to right and the size \(N\) of \(\trainset\) decreases from top to bottom.
What we learn from these graphs is that when going right and down, so when epistemic uncertainty increases, the gap between the robustness measures and uncertainty measures becomes larger.
This trend is furthermore the most prominent from left to right, so when the distribution shift is increased.

\begin{table}
    \centering
    \caption{The AU-ARCs for the NBC for all datasets with training set size \(N=50\) and feature noise \(\noise=0.2\).}\label{tab:auc_all_nbc_ds}
    \includegraphics[width=\linewidth]{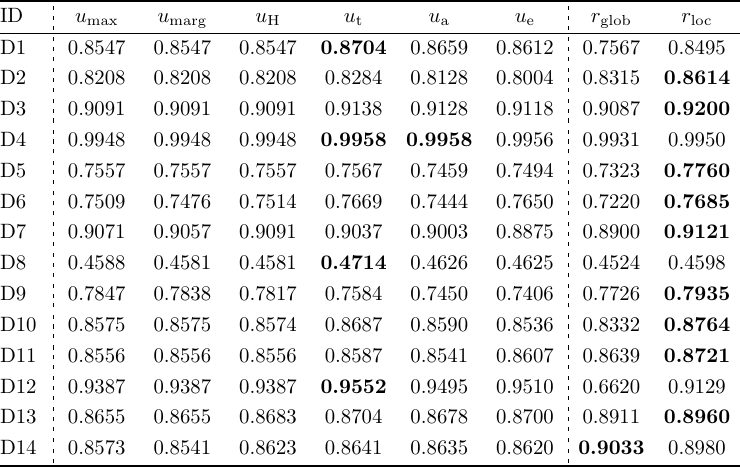}
\end{table}

To see to what extent this conclusion extends to other datasets, Table~\ref{tab:auc_all_nbc_ds} displays the (averaged) AU-ARCs for all considered datasets for the NBC, but now in the high uncertainty setting.
All values in this table are for \(N=50\) and \(\noise=0.20\).
In this setting, local robustness wins 9 out of 14 times, which is a clear increase over the 6 times we saw earlier in the standard setting. Global robustness is still not competitive, but it does win once now, and its perfomance seems to decrease a bit less than that of the uncertainty measures if we compare the standard setting to that with increased epistemic uncertainty.
These experiments thus confirm the behavior we expected: in the presence of epistemic uncertainty, robustness quantification (especially the local version) performs relatively better than uncertainty quantification.
That said, there are also some datasets where this trend is not noticeable, which motivates a further investigation of why and when robustness is useful compared to uncertainty.

\section{Combining Robustness and Uncertainty}\label{sec:combination}
\begin{figure}[!htb]
  \centering
  \includegraphics[width=0.5\linewidth]{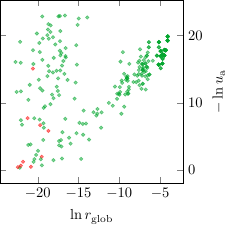}
  \caption{
	This point cloud (logarithmic scale) depicts for each instance if its predicted class was correct (green) or wrong (red). This plot was made for the dataset D4, performing classification with the NBC.}\label{fig:pointcloud}
\end{figure}

As has become clear in the previous section, it seems that both uncertainty and robustness measures are capable of assessing the reliability of predictions.
Furthermore, which of the approaches is better seems to depend on the particular dataset and model used, so both approaches seem to have their own strengths and weaknesses.
For that reason, we now proceed to investigate whether we can combine both types of measures to arrive at an even better reliability assessment.
To understand why this might indeed be possible, we take a look at the point cloud in Figure~\ref{fig:pointcloud}.
This point cloud represents each instance of \(\testset\) for dataset D4 with a colored dot, where green means that the instance was classified correctly, and red otherwise.
The \(x\)- and \(y\)-coordinate of a dot respectively represent the values for \(\robglob\) and \(\aleatunc\) on a logarithmic scale.
Since the points are spread over the plane, it means that for a given value of one of the measures, the other measure could be used to further distinguish the more and less reliable instances.
We also clearly see that misclassified instances tend to have both high uncertainty and low robustness (since the red dots are in the bottom left region).
This indicates that combining the two measures could indeed lead to an even better reliability assessment.

Given the complementary behavior of uncertainty and robustness measures, it seems logical to construct a hybrid reliability measure as a function that maps two numerical values, being the measures we'd like to combine, to a new numerical reliability value that performs even better at ordering instances than either of the two on their own.
It is not obvious, however, how to meaningfully combine two numerical values that capture different concepts of reliability into a single value.
Since we only aim to order the instances at this point, we therefore omit the step of constructing a hybrid measure, but instead directly aim to obtain a hybrid order of the instances.


To obtain an order that combines uncertainty and robustness, we take a weighted average of the two orders.
First, we order all instances with both measures separately to obtain for each instance two numbers that correspond to its position in each of the orders.
If for the \(i\)-th instance the position according to an uncertainty measure $u$ is \(n_{u, i}\), and according to a robustness measure \(r\) is \(n_{r, i}\), we determine the hybrid position of this instance using the weighted average of the two separate positions. In particular, we let
\[
    h_i \coloneq \gamma n_{u,i} + (1-\gamma)n_{r,i},
\]
where the weighting coefficient \(\gamma \in [0,1]\) determines the relative importance of uncertainty and robustness, and then order all instances in order of increasing \(h_i\), where ties are decided by the uncertainty measure.
In particular, \(\gamma=1\) leads to the same order as induced by uncertainty alone, and \(\gamma=0\) to the one for robustness.

Since we've already observed that the relative performance of robustness and uncertainty depends on the dataset, it is clear that the weighting coefficient \(\gamma\) should also depend on the dataset.
We therefore choose to optimize \(\gamma\) on the training set.
To do so, we compute the AU-ARC for a grid of possible values for \(\gamma\) and determine the \(\gamma\) that yields the highest AU-ARC for the training set, which we denote by \(\gamma_\mathrm{train}\). This is not the value of $\gamma$ that we'll use for the test set, though, since this would suffer from overfitting. To avoid this, we also consider a hyperparameter \(\mu\) that introduces a bias, using the following formula to obtain a biased verison \(\gamma^*\) of $\gamma_\mathrm{train}$:
\begin{equation}
    \gamma^*(\mu) = \begin{cases}
        (1-\mu)\gamma_\mathrm{train} + \mu &\text{if }\mu >0, \\
        (1+\mu)\gamma_\mathrm{train} &\text{else}.
    \end{cases}
\end{equation}
The hyperparameter \(\mu\) is obtained by doing 5-fold cross validation on the training set, optimizing for the best average AU-ARC across the 5 folds, and this value of $\mu$ is then used to calculate the \(\gamma^*\) that we use for the test set.
To even better protect ourselves against overfitting, we could of course also have tuned $\mu$ on a held-out validation set that is distinct from the training set.
However, we prefer for a reliability measure to provide extra information as an add-on to a model, without interfering with its performance as would be the case if we were to hold out data for the purpose of optimizing $\mu$.

Table~\ref{tab:AUC_hybrid_table_NBC} and Table~\ref{tab:AUC_hybrid_table_GeFs} show the resulting AU-ARCs for the NBC and GeFs, respectively, in the standard setting.
The first column contains the AU-ARC of the uncertainty measure we chose to combine with the robustness measures.
We chose \(\aleatunc\) for the NBC and \(\totunc\) for the GeFs, since these seemed to do best for the uncertainty measures based on the tables in Section~\ref{sec:comparison} (both won 3 times).
The third and seventh columns contain the AU-ARCs of the robustness measures with their corresponding hybrid results next to them.
The remaining columns provide the trained weighting coefficient \(\gamma^*\) used for combining the two measures, and the optimal \(\gamma_{\mathrm{opt}}\) that would have led to the highest possible AU-ARCs given these two measures (which we cannot learn without acces to the test set, but provides a nice reference point).
To make the results more easily interpretable, we highlighted in each row the AU-ARC of the hybrid measure whenever its value was the highest.
In most cases, the combination of uncertainty and robustness wins (indicated in bold).
Especially in the case of the NBC, combining the global robustness measure with uncertainty seems to work very well, which is surprising since the global robustness measure itself dit not perform all that well for the NBC.
The results for the hybrid approach look very nice for the GeFs as well. However, the values for \(\gamma^*\) are very polarized here.
Even though this does not seem to affect the performance too much, it is not ideal since it might indicate overfitting.
This becomes especially clear if we compare the found \(\gamma^*\) with the optimal one $\gamma_{\mathrm{opt}}$.\\

\begin{table}
    \centering
    \caption{
        The AU-ARCs for the NBC of \(\aleatunc\), \(\robglob\), \(\robloc\) and their combinations.
	    The columns next to each hybrid show the found \(\gamma^*\) and the optimal \(\gamma_{\mathrm{opt}}\) (on the test set).}\label{tab:AUC_hybrid_table_NBC}
    \includegraphics[width=\linewidth]{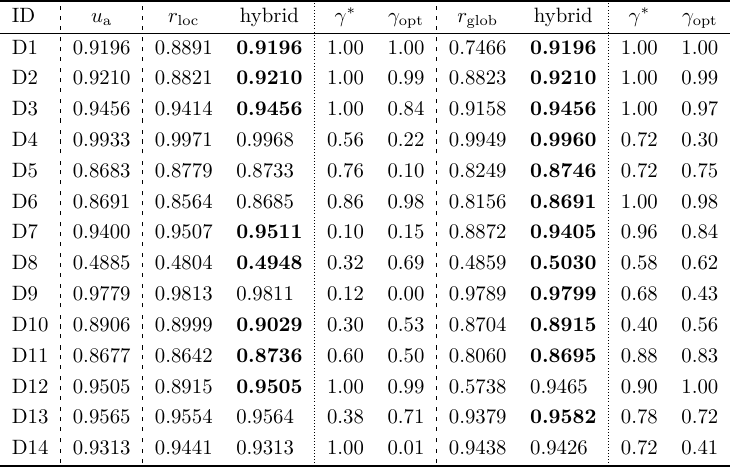}
\end{table}

\begin{table}
    \centering
    \caption{
        The AU-ARCs for the GeFs of \(\totunc\), \(\robglob\), \(\robloc\) and their combinations.
	    The columns next to each hybrid show the found \(\gamma^*\) and the optimal \(\gamma_{\mathrm{opt}}\) (on the test set).}\label{tab:AUC_hybrid_table_GeFs}
    \includegraphics[width=\linewidth]{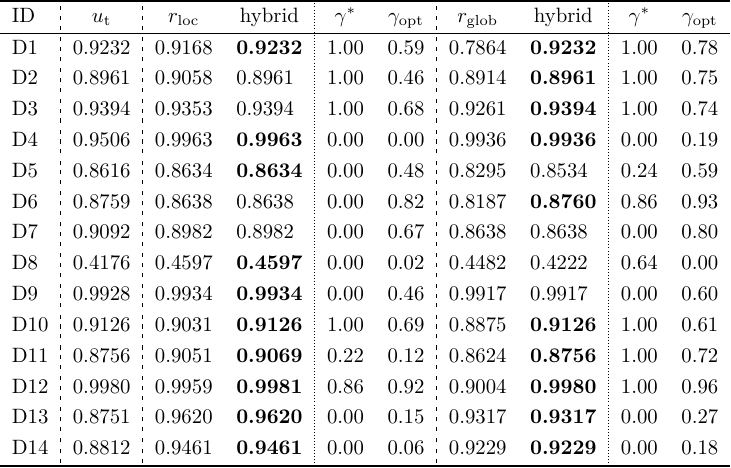}
\end{table}

We end by observing that, in addition to better reliability assessments, our hybrid approach also provides us with information about the relative importance of uncertainty and robustness for each dataset, in the form of the trained weighting coefficient \(\gamma^*\) (or the ideal one $\gamma_{\mathrm{opt}}$).
As can be seen from our results, this relative importance varies substantially between the datasets, and furthermore depends on the type of uncertainty and robustness that is considered.

\section{Discussion}
The take-away message of this contribution, in our view, is that uncertainty quantification and robustness quantification seem to be fundamentally different: both conceptually and in practice they each seem to have their own way of contributing to assessing the reliability of the predictions of a classifier. As our results clearly demonstrate, it is therefore worthwile to consider both approaches, and combine them to arrive at even better reliability assessments. There is, however, still much to explore in this regard, and we hope that our work can serve as a stepping stone for future research on this topic.


Regarding robustness quantification itself, we observed that the local robustness measures seem to outperform the global one in most cases, and that these local ones are often the best performing reliability measures overall, even outperforming the best uncertainty measures in several cases, especially so in the presence of distribution shift.
Inspired by these observations, it seems worthwile to try to expand the idea of local robustness quantification to more complex model architectures. If successful, this would allow the use of robustness quantification for image classification, which seems promising since that is a task in which distribution shift is very common.

\begin{acknowledgements}
    The work of both authors was partially supported by Ghent University's Special Research Fund, through Jasper De Bock's starting grant number 01N04819.
\end{acknowledgements}

\bibliography{detavernier_uai26}







\end{document}